\documentclass[a4paper,10pt]{article}
\usepackage[a4paper,width=150mm,top=25mm,bottom=25mm,lmargin=2.5cm,rmargin=2.5cm]{geometry}
\usepackage[utf8]{inputenc}
\usepackage{amsfonts}
\usepackage{amsmath}
\usepackage{graphicx}
\usepackage{caption}
\usepackage{subcaption}
\usepackage{booktabs}
\usepackage{fancyvrb}
\usepackage{xcolor}
\usepackage{multirow}
\usepackage{setspace}
\usepackage[ruled,vlined]{algorithm2e}
\usepackage[title]{appendix}
\graphicspath{ {figures/} }
\usepackage{fancyhdr}% http://ctan.org/pkg/fancyhdr

\fancypagestyle{conferencefooter}{%
  \fancyhf{}
  
  \fancyfoot[L]{\small 35th Conference on Neural Information Processing Systems (NeurIPS 2021), \textit{Strategic ML Workshop}}
}

\DeclareMathOperator*{\argmax}{argmax}
\DeclareMathOperator*{\argmin}{argmin}

\title{\vspace{-0.5cm}Improving Robustness of Malware Classifiers using Adversarial Strings Generated from Perturbed Latent Representations\vspace{0.2cm}}
\author{
    \textbf{Marek Galovic}\footnote{Department of Computer Science, Faculty of Electrical Engineering, Czech Technical University in Prague} \\ \texttt{\small galovic.galovic@gmail.com}
    \and
    \textbf{Branislav Bosansky}\footnote{Avast Software s.r.o} \footnotemark[1] \\ \texttt{\small branislav.bosansky@avast.com}
    \and
    \textbf{Viliam Lisy}\footnotemark[2] \footnotemark[1] \\ \texttt{\small viliam.lisy@avast.com}
}
\date{}

\begin{document}
\maketitle
\thispagestyle{conferencefooter}
\begin{abstract}
In malware behavioral analysis, the list of accessed and created files very often indicates whether the examined file is malicious or benign. However, malware authors are trying to avoid detection by generating random filenames and/or modifying used filenames with new versions of the malware. These changes represent real-world adversarial examples. The goal of this work is to generate realistic adversarial examples and improve the classifier's robustness against these attacks. Our approach learns latent representations of input strings in an unsupervised fashion and uses gradient-based adversarial attack methods in the latent domain to generate adversarial examples in the input domain. We use these examples to improve the classifier's robustness by training on the generated adversarial set of strings. Compared to classifiers trained only on perturbed latent vectors, our approach produces classifiers that are significantly more robust without a large trade-off in standard accuracy.
\end{abstract}
\section{Introduction}
Robustness of classifiers against adversarial attacks~\cite{goodfellow2015fgsm, papernot2016transferability, madry2019deep} is particularly relevant in security sensitive domains. We consider the problem of determining whether an executable application is benign or malicious based on the set of files the application accessed/created during runtime \cite{STIBOREK2018346}. Malware authors avoid detection by generating random filenames and/or perturbing existing ones. These changes represent real-world adversarial examples against the detection classifier. The success of attacks using these adversarial examples depends primarily on how much knowledge the adversary has about the target classifier and on the robustness of the classifier against adversarial examples. We focus on the latter part and aim to develop methods for training string classifiers that are robust against adversarial examples.
\begin{figure}[h]
    \centering
    \includegraphics[width=0.6\textwidth]{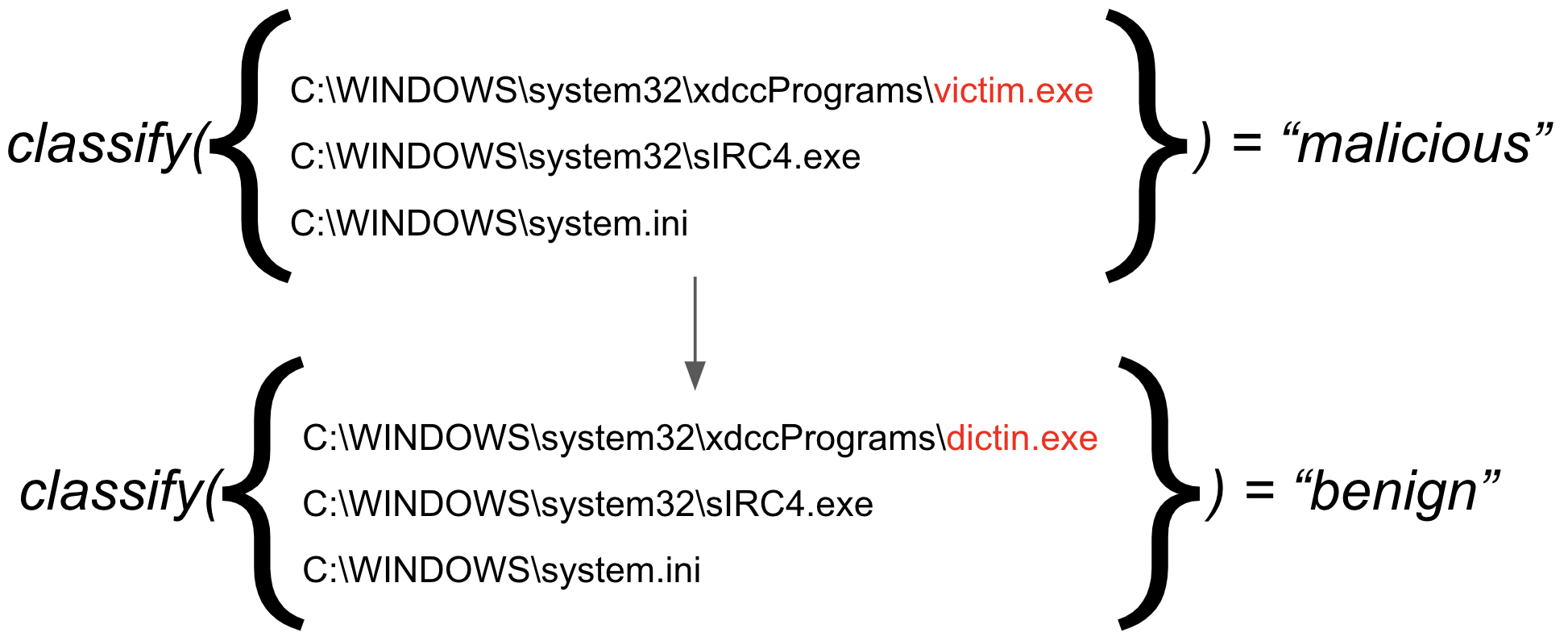}
    \caption{Desired Adversarial Perturbation}
    \label{fig:desired_adv_perturbation}
\end{figure}

We consider the following problem (see Figure~\ref{fig:desired_adv_perturbation}): Assume a classifier that classifies a set of strings representing accessed files as benign or malicious. What is the minimal sequence of perturbations that results in a given set of strings being misclassified? Unlike the image domain, any change in the set of strings will be perceptible by humans. Therefore, we are interested in perturbations that produce adversarial strings which are close to the original strings. In theory, we are guaranteed to find a minimal sequence of adversarial perturbations using combinatorial search algorithms \cite{Zang_2020, pmlr-v97-moon19a}. However, these methods are not feasible for our problem as the cardinality of the search space is $\approx 256^{nl}$, where $n$ is the number of strings in the set and $l$ is the average length of strings. Instead, we focus on adversarial perturbations in the continuous latent representation space.
% To make adversarial sample generation practically feasible in the domain of character sequences, 
As we have full access to the classifier, we can leverage its gradients to compute an adversarial perturbation in the latent representation space and then directly generate adversarial strings from the perturbed latent representation.

To make the generation of actual adversarial strings possible, we use an encoder-decoder style neural network \cite{graves2014generating} to learn latent representations of strings in an unsupervised fashion. We develop a convolutional-recurrent architecture that achieves high reconstruction accuracy on highly irregular/random strings\footnote{Note that strings corresponding to accessed files contain various special characters and/or numbers and thus differ significantly from standard words from natural language.}. This is important to us because we want generated changes to be caused by latent perturbations rather than high bias of the decoder. Using a pre-trained representation model, we then train a multiple instance classifier to classify sets of strings representing accessed files as benign or malicious.

Next, we use existing gradient-based methods to compute adversarial perturbations of latent string representations. We then generate adversarial strings using the decoder part of our network, where we use the perturbed latent representations as the decoder's initial state. This allows us to generate adversarial examples inside the training loop of the classifier and use them as additional input examples.

Finally, we explore the benefits of our method compared to adversarial training on perturbed latent vectors only (i.e., without generating actual adversarial sets of strings). We show that adversarial training on perturbed strings allows us to train classifiers that are significantly more robust without a large trade-off in standard accuracy.
Our contributions are as follows:
\begin{itemize}
    \item {We develop an encoder-decoder architecture that achieves high reconstruction accuracy on highly irregular/random strings.}
    \item {We generate realistic looking adversarial strings using gradient-based attack methods in the latent representation space.}
    \item {We compare the performance of classifiers trained on adversarial strings vs. classifiers trained only on perturbed latent vectors. We show that string adversarial training produces classifiers which are significantly more robust against adversarial examples without requiring a large trade-off in standard accuracy.}
\end{itemize}
\section{Related Work}
Adversarial perturbations for neural networks were first introduced by Szegedy et al. \cite{szegedy2014adv} and later explored by Goodfellow et al. \cite{goodfellow2015fgsm} in the domain of computer vision. They describe perturbations of input images that are imperceptible by humans but cause the image classifier to assign an incorrect label for the perturbed input. More importantly, different models trained on different subsets of data can misclassify the same adversarial example \cite{szegedy2014adv}. This suggests that it is possible to generate adversarial examples using a surrogate model to which the attacker has full access, and then use these adversarial examples to attack an unknown model of interest.

Adversarial attacks were also explored, among others, in the field of natural language processing which is relevant to ours because it operates on string inputs. Main approaches for adversarial attacks in the natural language processing domain can be categorized as sentence-level attacks \cite{zhao2018generating, wang2018robust}, word-level attacks \cite{samanta2017crafting,sato2018interpretable,zhang2020generating}, and character-level attacks \cite{ebrahimi-etal-2018-adversarial,gaoetal2018}. Sentence-level and word-level attacks seem to exploit high-level semantic information and often make use of synonyms or sentences with the same meaning as adversarial perturbations. We instead focus on character-level attacks which allows us to generate adversarial perturbations inside of individual file/directory names.

In this work, we consider input examples that are represented as sets of strings. Formally, this is known as Multiple Instance Learning (MIL) \cite{CARBONNEAU2018329}. Prior work describes three main paradigms for solving multiple instance learning problems, namely, \textit{instance-space paradigm}, \textit{bag-space paradigm} and \textit{embedding-space paradigm}. We follow the \textit{embedding-space paradigm} \cite{Chen2006MILEmbedding, Cheplygina_2015} as we first embed individual strings into vectors, and then aggregate instance vector representations into a fixed-size bag vector representation. Our bag aggregation function is similar to methods proposed in \cite{DBLP:journals/corr/abs-1802-04712, DBLP:journals/corr/abs-1810-00825}.

Existing methods for malware classification predict either a binary class (benign/malicious) \cite{Pascanu_2015_7178304, STIBOREK2018346} or a malware family \cite{DBLP:journals/corr/abs-1811-07842}. Furthermore, they can use hand engineered features, end-to-end learned feature extractors \cite{Pascanu_2015_7178304, Kalash_2018_8328749}, or a combination or both \cite{GIBERT2020101873}. Prior work also explores adversarial examples for malware classifiers. Grosse et al. \cite{grosse2016adversarial} explore gradient-based attack on classifiers that use binary feature vectors. Hu et al. \cite{hu2017generating} use Generative Adversarial Networks (GANs) to generate adversarial attacks without access to the target classifier. Kolosnjaji et al. \cite{8553214} propose a gradient-based method to adversarial attacks which changes only a few specific bytes at the end of malware sample.
\section{Methods}
\subsection{Vector Representations of Character Sequences}
\label{sec:encoder_decoder}
We aim to represent variable-length character sequences as fixed-size vectors in $\mathbb{R}^d$. We learn the latent representation of sequences from data in an unsupervised fashion. Firstly, we embed individual ASCII characters as vectors in $e_i \in \mathbb{R}^l;\ i=1...n$, where $l$ is the embedding size and $n$ is the sequence length. Secondly, we use an encoder function $F:\mathbb{R}^{n \times l} \to \mathbb{R}^d$ to represent a sequence of character embeddings as a vector in $\mathbb{R}^d$, where $n$ is the sequence length, $l$ is the embedding size, and $d$ is the latent representation size. Lastly, we use a decoder function $G:\mathbb{R}^d \to \mathbb{R}^{n \times 256}$ to reconstruct the original input sequence from its respective latent representation $h_T \in \mathbb{R}^d$. We normalize each of the decoder outputs $\hat{y}_t \in \mathbb{R}^{256};\ t=1...n$ using the $softmax$ function to obtain a valid probability distribution over ASCII characters at each output step $t$. Unlike the encoder, the decoder does not have access to the input sequence, but instead uses its outputs at step $t-1$ as inputs at step $t$. Hence, it autoregressively decodes the output sequence using only information encoded in the fixed-size vector representation $h_T$.

\begin{figure}[h]
\centering
\includegraphics[width=0.6\textwidth]{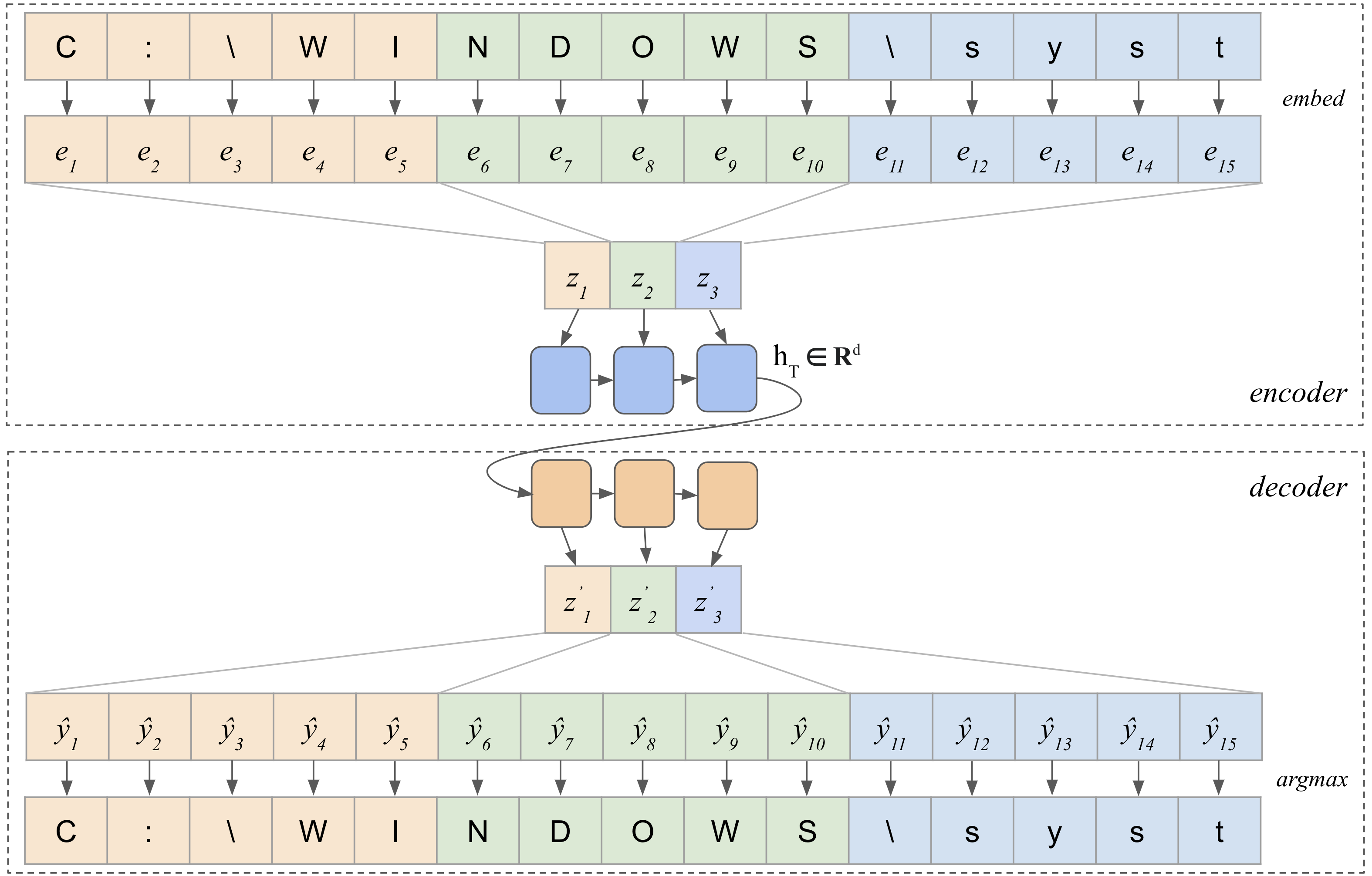}
\caption{Convolutional Recurrent Sequence-to-Sequence Autoencoder}
\label{conv_rnn_seq2seq}
\end{figure}

We use a combination of convolutional and recurrent neural networks for the encoder $F$ and the decoder $G$ (Figure \ref{conv_rnn_seq2seq}). The encoder first applies a 1D convolution with $kernelWidth=5$ and $stride=5$ to represent 5-grams as vectors and to reduce the length of the input to the recurrent layer. A recurrent neural network is then used to summarize the sequence of 5-gram vector representations into a single, fixed-size, vector representation. Given this vector representation, we then use an autoregressive decoder $G$ to obtain a reduced-length sequence of vectors and apply a transposed 1D convolution to obtain the output sequence with the same length as the input sequence.

\subsection{Multiple Instance Classification}
\label{sec:methods_mil}
Our approach to multiple instance classification follows the \textit{embedding-space paradigm} \cite{Chen2006MILEmbedding, Cheplygina_2015} and assumes that the labels are only available on the bag level. We use a transformation function $\mathcal{A}: \mathbb{R}^{k \times m} \to \mathbb{R}^{hd}$, where $k$ is the number of instances in the bag, $m$ is the size of vectors that represent instances, $h$ is the number of attention heads, and $d$ is the hidden size. Input to $\mathcal{A}$ is a matrix $E \in \mathbb{R}^{k \times m}$, where rows of the matrix are latent representations of character sequences. Output of $\mathcal{A}$ represents an aggregated fixed-size vector representation of the input bag.

\begin{equation}
\label{eqn:lba_keys}
K := EW_K \in \mathbb{R}^{k \times d}
\end{equation}
\begin{equation}
\label{eqn:lba_values}
V := EW_V \in \mathbb{R}^{k \times d}
\end{equation}
\begin{equation}
\label{eqn:lba_weights}
W := W_QK^T \in \mathbb{R}^{h \times k}
\end{equation}
\begin{equation}
\label{eqn:lba_weights_norm}
w_i = \frac{exp(w_i)}{\sum_{j=1}^{k}exp(w_j)}; i=1...k
\end{equation}
\begin{equation}
\label{eqn:lba_out}
o := vec(WV) \in \mathbb{R}^{hd}
\end{equation}

Our choice of the aggregation function $\mathcal{A}$ is similar to a transformer layer \cite{vaswani2017attention, DBLP:journals/corr/abs-1802-04712}. It has three trainable matrices: key projection matrix $W_K \in \mathbb{R}^{m \times d}$, value projection matrix $W_V \in \mathbb{R}^{m \times d}$, and a query matrix $W_Q \in \mathbb{R}^{h \times d}$, where $h$ is the number of attention heads \cite{vaswani2017attention} and $d$ is the hidden size. Inputs $E$ are first projected to keys and values as described by Equation \ref{eqn:lba_keys} and Equation \ref{eqn:lba_values}. Multiplying keys with the query matrix $W_Q$ (Equation \ref{eqn:lba_weights}) gives us a matrix of attention weights whose columns are then $\textit{softmax}$ normalized to unit sum (Equation \ref{eqn:lba_weights_norm}). Finally, multiplying the normalized weights matrix $W$ with the values matrix $V$ produces a matrix of convex combinations of value vectors for each head. This matrix is then flattened to a vector of fixed dimension $\mathbb{R}^{hd}$ (Equation \ref{eqn:lba_out}) which represents the aggregated output. The aggregated bag-level representation is used as input features to a classifier which is a simple feed-forward neural network with two output classes (malicious/benign). Using this setup, we obtain a differentiable classifier function $f_{\theta}:\mathbb{R}^{k \times m} \to \mathbb{R}^2$ that allows us to compute adversarial perturbations of instance vectors using gradient-based methods.

\subsection{Adversarial Sample Generation}
\label{sec:adv_sample_generation}
We now describe our approach for generating adversarial examples using gradient-based perturbations of latent representations. Given a trained encoder $F$, decoder $G$ and a multiple instance classifier $C$, we generate adversarial samples in the following way. First, using the encoder $F$ we encode all instances $S = \{s_1, ..., s_n\}$ in the bag to their respective latent representations as $Z = \{z_i = F(s_i);\ \forall z_i \in \mathbb{R}^d,\ i\in\{1,...,n\}\};\ Z \in \mathbb{R}^{n \times d}$, where $n$ is the number of instances and $d$ is the latent representation size. Then, we use the multiple instance classifier $C$ to obtain a perturbation of these latent representations as $\delta^{\ast} = \argmax_{\delta \in \Delta}\ell(C(Z + \delta),y);\ \delta^{\ast} \in \mathbb{R}^{n \times d}$. Finally, we generate a set of adversarial instances in the original (string) domain as $S_{adv} = \{s^{adv}_i = G(Z_i + \delta^{\ast}_i);\ i \in \{1,...,n\}\}$, where $Z_i \in \mathbb{R}^d$ and $\delta^{\ast}_i \in \mathbb{R}^d$.

We modify existing gradient-based attack methods, namely Projected Gradient Descent (PGD) (Algorithm \ref{alg:pgd}) and Fast Gradient Sign Method (FGSM) (Algorithm \ref{alg:fgsm}), to approximately find an optimal perturbation $\delta^{\ast} \in \Delta$, where $\Delta$ is a set of allowable perturbations. The modifications are due to the fact that the encoder $F$ and the decoder $G$ are not a perfect inverse of each other, i.e. it does not necessarily hold that $F(G(z)) = z$ for some latent representation $z \in \mathbb{R}^d$. Because of this, we may find a perturbation $\delta^{\ast}$ that is adversarial in the latent representation space, i.e. $C(z + \delta^{\ast}) \neq y$, but the bag generated from the perturbed state $z + \delta^{\ast}$ is not adversarial, i.e. $C(F(G(z + \delta^{\ast}))) = y$. Therefore, our modified algorithm generates an adversarial bag after each gradient step, encodes this generated bag to its respective latent representation, classifies the latent representation, and checks if the label is different from the true label. Only misclassified bags generated from perturbed latent states are considered to be successful adversarial examples.
\BlankLine
\noindent
\begin{minipage}{0.48\textwidth}
    \begin{algorithm}[H]
    \footnotesize
    \label{alg:pgd}
    \SetAlgoLined
    \KwIn{$paths,y,\alpha,\mathcal{P},T$}
    \KwOut{$advPaths$}
    $state \leftarrow encode(paths)$\;
    $\delta_0 \leftarrow zerosLike(state)$\;
    $\hat{y} \leftarrow classify(state)$\;
    \If{$\hat{y} \neq y$} {
        \Return $paths$\;
    }
    \For{i = 1...T} {
        $\nabla_{\delta} \leftarrow \nabla_{\delta}\ell(classify(state + \delta_{i-1}), y)$\;
        $\delta_i \leftarrow \mathcal{P}(\delta_{i-1} + \alpha\frac{\nabla_{\delta}}{||\nabla_{\delta}||_2 + \gamma})$\;
        $advPaths \leftarrow decode(state + \delta_i)$\;
        $advState \leftarrow encode(advPaths)$\;
        $\hat{y} \leftarrow classify(advState)$\;
        \If{$\hat{y} \neq y$} {
            \Return $advPaths$\;
        }
    }
    \caption{Modified PGD}
    \end{algorithm}
\end{minipage}
\hfill
\begin{minipage}{0.48\textwidth}
    \begin{algorithm}[H]
    \footnotesize
    \label{alg:fgsm}
    \SetAlgoLined
    \KwIn{$paths,y,\epsilon,\epsilon_{max},\delta_{\epsilon}$}
    \KwOut{$advPaths$}
    $state \leftarrow encode(paths)$\;
    $\hat{y} \leftarrow classify(state)$\;
    \If{$\hat{y} \neq y$} {
        \Return $paths$\;
    }
    $\nabla_{state} \leftarrow \nabla_{state}\ell(classify(state), y)$\;
    \While{$\epsilon \leq \epsilon_{max}$} {
        $advPaths \leftarrow decode(state + \epsilon sgn(\nabla_{state}))$\;
        $advState \leftarrow encode(advPaths)$\;
        $\hat{y} \leftarrow classify(advState)$\;
        \If{$\hat{y} \neq y$} {
            \Return $advPaths$\;
        }
        $\epsilon \leftarrow \epsilon + \delta_{\epsilon}$\;
    }
    \caption{Modified FGSM}
    \end{algorithm}
\end{minipage}

\subsection{Robust Classifier Training}
To train a classifier that is robust to adversarial examples, we augment the training dataset with adversarial examples generated using methods described in the previous section (Algorithm \ref{alg:adversarial_training}). Given a classifier function $f_{\theta}$ with parameters $\theta$, we modify the training procedure from a simple minimization of the classification loss $\ell(f_{\theta}(x), y)$ to minimax optimization (Equation \ref{eqn:adv_minimax_objective}). The outer minimization tries to minimize the classification loss on the training set $\mathcal{T}$, while the inner maximization tries to maximize it by adversarially perturbing the input.
\begin{equation}
\label{eqn:adv_minimax_objective}
\theta^{\ast} = \argmin_{\theta}\frac{1}{|\mathcal{T}|}\sum_{i=1}^{|\mathcal{T}|} \max_{\delta \in \Delta}\ell(f_{\theta}(x_i + \delta), y_i)
\end{equation}
The update rule for classifier's parameters $\theta$ is then defined as Equation \ref{eqn:minimax_update}.
\begin{equation}
\label{eqn:minimax_update}
\theta := \theta - \alpha\nabla_{\theta}\max_{\delta \in \Delta}\ell(f_{\theta}(x + \delta), y)
\end{equation}

To compute gradient of the inner maximization we make use of the Danskin’s Theorem, which states that the gradient of the $max$ operator is equal to the gradient of the inner function evaluated at the maximum point (Equation \ref{eqn:adv_danskin}).
\begin{equation}
\label{eqn:adv_danskin}
\nabla_{\theta}\max_{\delta \in \Delta}\ell(f_{\theta}(x + \delta), y) = \nabla_{\theta}\ell(f_{\theta}(x + \delta^{\ast}), y)
\end{equation}
However, this result only applies for the case where we can compute the maximum exactly and that maximum is unique, neither of which can be guaranteed for our case as we are only able to solve the maximization approximately. Nevertheless, the stronger the adversarial attack is, the better approximation of the true maximum we get, and thus we also get a better approximation of the gradient.
\BlankLine
\BlankLine
\begin{algorithm}[H]
\footnotesize
\label{alg:adversarial_training}
\SetAlgoLined
\For{$\mathcal{B} \subseteq \mathcal{T}$} {
    $\nabla_{\theta} \leftarrow \textbf{0}$\;
    \For{$(x,y) \in \mathcal{B}$} {
        $\nabla_{\theta} \leftarrow \nabla_{\theta} + \nabla_{\theta}\ell(f_{\theta}(x), y)$\;
        \If{$f_{\theta}(x) = y$}{
            $\delta^{\ast} \leftarrow \argmax_{\delta \in \Delta}\ell(f_{\theta}(x + \delta), y)$\;
            $\nabla_{\theta} \leftarrow \nabla_{\theta} + \nabla_{\theta}\ell(f_{\theta}(x + \delta^{\ast}), y)$\;
        }
    }
    $\theta \leftarrow \theta - \frac{\alpha}{|\mathcal{B}|}\nabla_{\theta}$\;
}
\caption{Adversarial Training}
\end{algorithm}
\section{Experiments}
In this section we describe our experimental setup and achieved results. 
Unfortunately, there is no public dataset on the classification of set of strings similar to the problem of malware classification and thus we use a proprietary dataset that is the result of a proprietary emulator of a major antivirus vendor. 
This dataset is split according to a timestamp into training and testing subsets. 
The temporal split is important as it captures future modifications of the malware that are not known at the training time. 

The encoder-decoder model (Section \ref{sec:encoder_decoder}) is pre-trained to represent individual strings as fixed-size vectors and to reconstruct input strings from their respective vector representations. We use $hiddenSize=128$ and $kernelWidth=5$ throughout our experiments. A full list of the encoder-decoder hyperparameters and their performance is provided in Appendix \ref{appendix:enc_dec_hyperparams}. We then freeze the model weights of both the encoder and the decoder, and use the encoder's latent representations to train a multiple instance classifiers (Section \ref{sec:methods_mil}). We set the number of attention heads to $heads=8$ and provide a full list of the classifier hyperparameters and their performance in Appendix \ref{appendix:clf_hyperparams}.

Using the data and the classifier, we first investigate the attack success rate and quality of generated adversarial examples for methods described in Section \ref{sec:adv_sample_generation} to find appropriate parameters for the algorithms that generate adversarial attacks. 
Second, we compare standard accuracy and adversarial robustness between the original (non-robust) classifier, robust classifier trained on latent perturbations, and robust classifier trained on string perturbations. 

\subsection{Quality of Generated Adversarial Samples}
\begin{figure}[!h]
    \centering
    \begin{subfigure}[b]{0.49\textwidth}
        \centering
        \includegraphics[width=\textwidth]{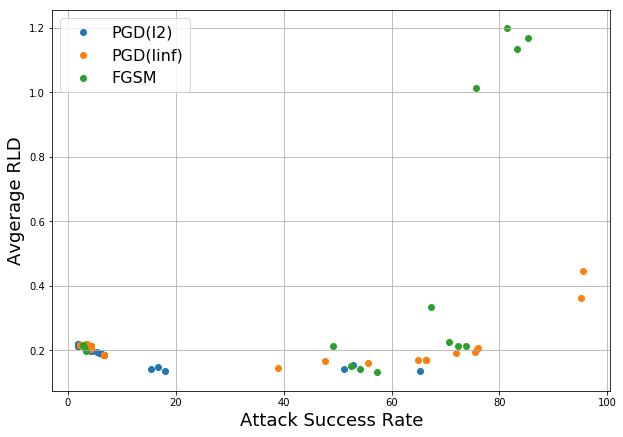}
        \caption{Performance of Attack Methods}
        \label{fig:adv_attack_solutions}
    \end{subfigure}
    \hfill
    \begin{subfigure}[b]{0.49\textwidth}
        \centering
        \includegraphics[width=\textwidth]{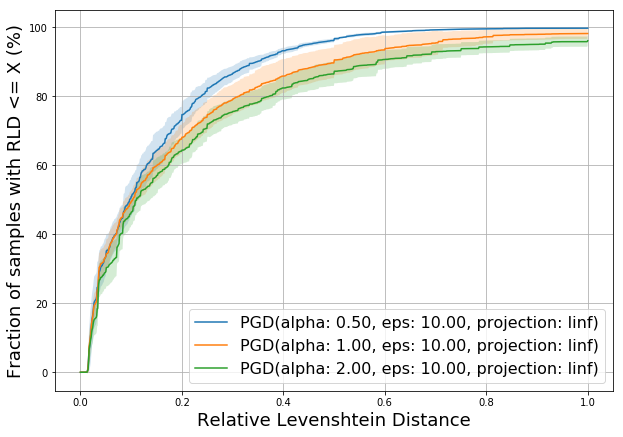}
        \caption{eCDF of Relative Levenshtein Distances}
        \label{fig:adv_path_similarity}
    \end{subfigure}
    \caption{Comparison of Adversarial Attack Methods}
    \label{fig:attack_methods_comparison}
\end{figure}

We compare the quality of generated adversarial strings across multiple gradient-based attack methods with different hyperparameters. Our goal is to find methods with high attack success rate that generate adversarial strings which are close to the original input. As a measure of string similarity, we define a Relative Levenshtein Distance (RLD) as $RLD = Levenshtein Distance\ /\ Input$ $Length$, where lower RLD means that strings are more similar. Figure \ref{fig:adv_attack_solutions} shows the adversarial attack success rate and average RLD between perturbed inputs and original inputs for both FGSM and PGD. We explore different hyperparameter settings for the projection operator $\mathcal{P}$, PGD step size $\alpha$, FGSM step size $\delta_{\epsilon}$, and $\epsilon$-neighborhood size $\epsilon$. Table \ref{tab:pareto_optimal_attack_methods} then shows attack success rate and average RLD for all Pareto optimal solutions, where adversarial strings were generated using 7 distinct classifiers to obtain uncertainty estimates. We use the Pareto optimal set of solutions to choose methods and their respective hyperparameters for further experiments. Additionally, we select methods with different adversarial attack success rates and compare the similarity of generated adversarial strings to the original inputs. Figure \ref{fig:adv_path_similarity} shows empirical cumulative distribution function of Relative Levenshtein Distances for attack methods that achieved the following average attack success rates: 63.86\% (blue), 74.42\% (orange) and 87.04\% (green).

\begin{table}[!h]
    \small
    \centering
    \begin{tabular}{lcc}
    \toprule
                                             Method & Attack Success Rate & Average RLD \\
    \midrule
        FGSM(delta: 0.01, max\_eps: 1.00)               &      56.13\% ± 2.23\% &  0.122 ± 0.017 \\
    PGD(alpha: 2.00, eps: 10.00, projection: l2)   &      62.67\% ± 3.45\% &  0.141 ± 0.015 \\
    PGD(alpha: 0.50, eps: 5.00, projection: linf)  &      63.83\% ± 3.69\% &  0.146 ± 0.015 \\
    PGD(alpha: 0.50, eps: 10.00, projection: linf) &      63.86\% ± 3.63\% &  0.146 ± 0.015 \\
    PGD(alpha: 1.00, eps: 2.00, projection: linf)  &      69.83\% ± 1.45\% &  0.162 ± 0.019 \\
    PGD(alpha: 1.00, eps: 5.00, projection: linf)  &      74.32\% ± 2.64\% &  0.193 ± 0.039 \\
    PGD(alpha: 1.00, eps: 10.00, projection: linf) &      74.42\% ± 2.56\% &  0.213 ± 0.057 \\
    PGD(alpha: 2.00, eps: 2.00, projection: linf)  &      76.91\% ± 3.92\% &  0.168 ± 0.021 \\
    PGD(alpha: 2.00, eps: 5.00, projection: linf)  &      86.76\% ± 4.43\% &  0.300 ± 0.066 \\
    PGD(alpha: 2.00, eps: 10.00, projection: linf) &      87.04\% ± 4.80\% &  0.371 ± 0.098 \\
    \bottomrule
    \end{tabular}
    \caption{Pareto Optimal Solutions}
    \label{tab:pareto_optimal_attack_methods}
\end{table}

These experiments show that it is possible to find methods with high attack success rate which also generate adversarial examples that are close to the original input. We compared FGSM and PGD with both $\ell_2$ and $\ell_{\infty}$ projections, where PGD with $\ell_{\infty}$ projection produced the best adversarial examples, FGSM produced adversarial examples with high attack success rate but with too many perturbations, and PGD with $\ell_2$ projection produced adversarial examples that are close to the original input but have low attack success rate. Therefore, we use PGD with $\ell_{\infty}$ projection for further experiments and vary the step size $\alpha$ to control the strength of adversarial attacks.

\subsection{Adversarial Training}
To demonstrate benefits of adversarial training on generated strings over perturbed latent vectors only, we first adversarially train classifiers using both methods, and then compare their robustness on adversarial examples generated using the original (non-robust) classifier, robust classifier trained on latent perturbations, and robust classifier trained on string perturbations.

% \begin{table}[!h]
%     \small
%     \centering
%     \begin{tabular}{clc||cc}
%     \toprule
%     \multicolumn{3}{l||}{} & \multicolumn{2}{c}{\textbf{Target Model Robustness}} \\
%     {} & {} & Standard Accuracy & Adversarial - Full & Adversarial - Latent \\
%     \midrule
    
%     \multirow{3}{*}{\rotatebox[origin=c]{90}{\parbox[c]{1.2cm}{\centering \textbf{Attack Model}}}} & Non-robust     &   \textbf{93.49\%} &  \textbf{94.49\%} &  90.44\% \\
%     {} & Adversarial - Full   &   92.05\% &  76.70\% &  79.03\% \\
%     {} & Adversarial - Latent &   91.90\% &  87.75\% &  26.38\% \\
%     \bottomrule
%     \end{tabular}
%     \caption{Standard Accuracy and Robustness of Adversarially Trained Classifiers}
%     \label{tab:classifiers_adv_robustness}
% \end{table}
\begin{table}[!h]
    \centering
    \begin{tabular}{lc}
    \toprule
    {Classifier} & Standard Accuracy \\
    \midrule
    
    Non-robust     &   \textbf{93.59\% ± 0.11\%} \\
    Adversarial - Latent &   91.84\% ± 0.13\% \\
    Adversarial - Full   &   91.18\% ± 0.85\% \\
    \bottomrule
    \end{tabular}
    \caption{Standard Accuracy of Adversarially Trained Classifiers}
    \label{tab:adv_classifiers_standard_acc}
\end{table}

\begin{table}[!h]
    \centering
    \begin{tabular}{lcc}
    \toprule
    {} & \multicolumn{2}{c}{\textbf{Target Model}} \\
    {\textbf{Attack Model}} & Adversarial - Latent & Adversarial - Full \\
    \midrule
    
    Non-robust & \ 89.95\% ± 1.09\% &  \textbf{95.20\% ± 1.42\%} \\
    Adversarial - Latent &  65.43\% ± 5.07\% &  90.95\% ± 1.80\% \\
    Adversarial - Full   &  76.84\% ± 2.25\% &  85.05\% ± 2.00\% \\
    \bottomrule
    \end{tabular}
    \caption{Robustness of Adversarially Trained Classifiers}
    \label{tab:adv_classifiers_robustness}
\end{table}

Table \ref{tab:adv_classifiers_standard_acc} and Table \ref{tab:adv_classifiers_robustness} show standard classification accuracy and adversarial attack success rates for all classifiers, where in Table \ref{tab:adv_classifiers_robustness} rows are models that generated adversarial attacks and columns are target models that were attacked. The reported results are obtained using 7 runs per classifier type. As expected, standard classification accuracy drops for both adversarially trained classifiers. However, the classifier trained on generated adversarial strings is significantly more robust compared to the classifier trained only on perturbed latent vectors. The classifier trained on adversarial strings is also more robust against adversarial attacks generated using the classifier trained only on perturbed latent vectors (90.95\% ± 1.80\% vs. 65.43\% ± 5.07\%). We conjecture that this is due to the fact that classifiers trained only on perturbed latent representations are less robust and therefore adversarial examples generated using these classifiers are weaker.

\begin{table}[!h]
    \centering
    \begin{tabular}{lcc}
    \toprule
            Classifier & Standard Accuracy & Robustness \\
    \midrule
        Full - $\alpha$=0.5 &            \textbf{92.19\%} &     94.62\% \\
          Full - $\alpha$=1 &            92.05\% &     94.51\% \\
     Latent - $\alpha$=0.05 &            91.90\% &     90.44\% \\
      Latent - $\alpha$=0.1 &            89.33\% &     93.64\% \\
          Full - $\alpha$=2 &            88.02\% &     \textbf{98.29\%} \\
      Latent - $\alpha$=0.2 &            82.21\% &     97.45\% \\
    \bottomrule
    \end{tabular}
    \caption{Standard Accuracy vs. Robustness}
    \label{tab:classifiers_adv_robustness_across_alpha}
\end{table}

Additionally, we train multiple robust classifiers using both methods with varying step size $\alpha$ to compare their robustness with regards to the drop in standard accuracy (Table \ref{tab:classifiers_adv_robustness_across_alpha}). In this experiment, we use adversarial examples generated using the original (non-robust) classifier for adversarial attacks. Robust classifiers trained on generated adversarial strings generally outperform robust classifiers trained only on perturbed latent vectors at comparable levels of standard classification accuracy. We omit uncertainty estimates due to computational resources needed to train a large number of adversarial classifiers.
\section{Conclusion}
In this work we have developed a method to efficiently generate adversarial strings in multiple instance learning setting. This allowed us perform adversarial training on perturbed sets of strings instead of perturbed latent representations. We experimented with standard (non-robust) classifiers, latent adversarial classifiers, and string adversarial classifiers where we compared their standard accuracy and robustness against adversarial examples. Classifiers trained on adversarial strings were the most robust against adversarial attacks and achieved similar standard accuracy as classifiers trained only on perturbed latent representations.

\small
\bibliographystyle{abbrv}
\bibliography{references}

\begin{thebibliography}{10}

\bibitem{DBLP:journals/corr/abs-1811-07842}
B.~Alsulami and S.~Mancoridis.
\newblock Behavioral malware classification using convolutional recurrent
  neural networks.
\newblock {\em CoRR}, abs/1811.07842, 2018.

\bibitem{CARBONNEAU2018329}
M.-A. Carbonneau, V.~Cheplygina, E.~Granger, and G.~Gagnon.
\newblock Multiple instance learning: A survey of problem characteristics and
  applications.
\newblock {\em Pattern Recognition}, 77:329--353, 2018.

\bibitem{Cheplygina_2015}
V.~Cheplygina, D.~M. Tax, and M.~Loog.
\newblock Multiple instance learning with bag dissimilarities.
\newblock {\em Pattern Recognition}, 48(1):264–275, Jan 2015.

\bibitem{ebrahimi-etal-2018-adversarial}
J.~Ebrahimi, D.~Lowd, and D.~Dou.
\newblock On adversarial examples for character-level neural machine
  translation.
\newblock In {\em Proceedings of the 27th International Conference on
  Computational Linguistics}, pages 653--663, Santa Fe, New Mexico, USA, Aug.
  2018. Association for Computational Linguistics.

\bibitem{gaoetal2018}
J.~{Gao}, J.~{Lanchantin}, M.~L. {Soffa}, and Y.~{Qi}.
\newblock Black-box generation of adversarial text sequences to evade deep
  learning classifiers.
\newblock In {\em 2018 IEEE Security and Privacy Workshops (SPW)}, pages
  50--56, 2018.

\bibitem{GIBERT2020101873}
D.~Gibert, C.~Mateu, and J.~Planes.
\newblock Hydra: A multimodal deep learning framework for malware
  classification.
\newblock {\em Computers and Security}, 95:101873, 2020.

\bibitem{goodfellow2015fgsm}
I.~J. Goodfellow, J.~Shlens, and C.~Szegedy.
\newblock Explaining and harnessing adversarial examples, 2015.

\bibitem{graves2014generating}
A.~Graves.
\newblock Generating sequences with recurrent neural networks, 2014.

\bibitem{grosse2016adversarial}
K.~Grosse, N.~Papernot, P.~Manoharan, M.~Backes, and P.~McDaniel.
\newblock Adversarial perturbations against deep neural networks for malware
  classification, 2016.

\bibitem{hu2017generating}
W.~Hu and Y.~Tan.
\newblock Generating adversarial malware examples for black-box attacks based
  on gan, 2017.

\bibitem{DBLP:journals/corr/abs-1802-04712}
M.~Ilse, J.~M. Tomczak, and M.~Welling.
\newblock Attention-based deep multiple instance learning.
\newblock {\em CoRR}, abs/1802.04712, 2018.

\bibitem{Kalash_2018_8328749}
M.~Kalash, M.~Rochan, N.~Mohammed, N.~D.~B. Bruce, Y.~Wang, and F.~Iqbal.
\newblock Malware classification with deep convolutional neural networks.
\newblock In {\em 2018 9th IFIP International Conference on New Technologies,
  Mobility and Security (NTMS)}, pages 1--5, 2018.

\bibitem{8553214}
B.~Kolosnjaji, A.~Demontis, B.~Biggio, D.~Maiorca, G.~Giacinto, C.~Eckert, and
  F.~Roli.
\newblock Adversarial malware binaries: Evading deep learning for malware
  detection in executables.
\newblock In {\em 2018 26th European Signal Processing Conference (EUSIPCO)},
  pages 533--537, 2018.

\bibitem{DBLP:journals/corr/abs-1810-00825}
J.~Lee, Y.~Lee, J.~Kim, A.~R. Kosiorek, S.~Choi, and Y.~W. Teh.
\newblock Set transformer.
\newblock {\em CoRR}, abs/1810.00825, 2018.

\bibitem{madry2019deep}
A.~Madry, A.~Makelov, L.~Schmidt, D.~Tsipras, and A.~Vladu.
\newblock Towards deep learning models resistant to adversarial attacks, 2019.

\bibitem{pmlr-v97-moon19a}
S.~Moon, G.~An, and H.~O. Song.
\newblock Parsimonious black-box adversarial attacks via efficient
  combinatorial optimization.
\newblock In K.~Chaudhuri and R.~Salakhutdinov, editors, {\em Proceedings of
  the 36th International Conference on Machine Learning}, volume~97 of {\em
  Proceedings of Machine Learning Research}, pages 4636--4645. PMLR, 09--15 Jun
  2019.

\bibitem{papernot2016transferability}
N.~Papernot, P.~McDaniel, and I.~Goodfellow.
\newblock Transferability in machine learning: from phenomena to black-box
  attacks using adversarial samples, 2016.

\bibitem{Pascanu_2015_7178304}
R.~Pascanu, J.~W. Stokes, H.~Sanossian, M.~Marinescu, and A.~Thomas.
\newblock Malware classification with recurrent networks.
\newblock In {\em 2015 IEEE International Conference on Acoustics, Speech and
  Signal Processing (ICASSP)}, pages 1916--1920, 2015.

\bibitem{samanta2017crafting}
S.~Samanta and S.~Mehta.
\newblock Towards crafting text adversarial samples, 2017.

\bibitem{sato2018interpretable}
M.~Sato, J.~Suzuki, H.~Shindo, and Y.~Matsumoto.
\newblock Interpretable adversarial perturbation in input embedding space for
  text, 2018.

\bibitem{STIBOREK2018346}
J.~Stiborek, T.~Pevny, and M.~Rehak.
\newblock Multiple instance learning for malware classification.
\newblock {\em Expert Systems with Applications}, 93:346--357, 2018.

\bibitem{szegedy2014adv}
C.~Szegedy, W.~Zaremba, I.~Sutskever, J.~Bruna, D.~Erhan, I.~Goodfellow, and
  R.~Fergus.
\newblock Intriguing properties of neural networks, 2014.

\bibitem{vaswani2017attention}
A.~Vaswani, N.~Shazeer, N.~Parmar, J.~Uszkoreit, L.~Jones, A.~N. Gomez,
  L.~Kaiser, and I.~Polosukhin.
\newblock Attention is all you need, 2017.

\bibitem{wang2018robust}
Y.~Wang and M.~Bansal.
\newblock Robust machine comprehension models via adversarial training, 2018.

\bibitem{Chen2006MILEmbedding}
{Yixin Chen}, {Jinbo Bi}, and J.~Z. {Wang}.
\newblock Miles: Multiple-instance learning via embedded instance selection.
\newblock {\em IEEE Transactions on Pattern Analysis and Machine Intelligence},
  28(12):1931--1947, 2006.

\bibitem{Zang_2020}
Y.~Zang, F.~Qi, C.~Yang, Z.~Liu, M.~Zhang, Q.~Liu, and M.~Sun.
\newblock Word-level textual adversarial attacking as combinatorial
  optimization.
\newblock {\em Proceedings of the 58th Annual Meeting of the Association for
  Computational Linguistics}, 2020.

\bibitem{zhang2020generating}
H.~Zhang, H.~Zhou, N.~Miao, and L.~Li.
\newblock Generating fluent adversarial examples for natural languages, 2020.

\bibitem{zhao2018generating}
Z.~Zhao, D.~Dua, and S.~Singh.
\newblock Generating natural adversarial examples, 2018.

\end{thebibliography}

\newpage
\normalsize
\appendix
\section{Adversarial Input Examples}
\label{appendix:adversarial_samples}
\begin{Verbatim}[commandchars=\&\{\}]
Input:
C:\Documents and Settings\Administrator\Application Data\Yandex\ui
Adversarial input:
C:\Documents and Settings\Administrator\Application Data\Yandex\&textcolor{red}{ote.exi}

Input:
C:\WINDOWS\Temp\GUM896.tmp\goopdateres_uk.dll
Adversarial input:
C:\WINDOWS\Temp\GUM896.tmp\goopdateres_&textcolor{red}{r}k.dll

Input:
C:\Program Files\GUMA36C.tmp\goopdateres_en-GB.dll
Adversarial input:
C:\Program Files\&textcolor{red}{E}UMA36C.tmp\goopdateres_en-&textcolor{red}{d1}.ddl&textcolor{red}{.exe}

Input:
C:\WINDOWS\Temp\oCFVhbs.ini
Adversarial input:
C:\WINDOWS\Temp\&textcolor{red}{U}C&textcolor{red}{Bsto}s.&textcolor{red}{e}ni

Input:
C:\WINDOWS\Temp\nsb7261.tmp\UAC.dll
Adversarial input:
C:\WINDOWS\Temp\ns&textcolor{red}{c653}1.tmp\UAC.dll

Input:
C:\WINDOWS\Temp\BgRh53b.ini
Adversarial input:
C:\WINDOWS\Temp\&textcolor{red}{1OnDoe}b&textcolor{red}{F.bae}

Input:
C:\WINDOWS\Temp\nsk6C58.tmp&textcolor{red}{\UAC.dll}
Adversarial input:
C:\WINDOWS\Temp\nsk6C58.tmp

Input:
C:\WINDOWS\Temp\i4j_nlog_2
Adversarial input:
C:\WINDOWS\Temp\i4j_nlog&textcolor{red}{dllgodb.g.dxe}

Input:
C:\WINDOWS\Temp\GUM896.tmp\goopdateres_zh-TW.dll
Adversarial input:
C:\WINDOWS\Temp\GUM896.tmp\goopdateres_zh-T&textcolor{red}{R}.dll

Input:
C:\WINDOWS\Temp\lkpYxWW.ini
Adversarial input:
C:\WINDOWS\Temp\&textcolor{red}{wL}pY&textcolor{red}{IzC.e}ni

Input:
C:\WINDOWS\Temp\nsg28B0.tmp\System.dll
Adversarial input:
C:\WINDOWS\Temp\nsg&textcolor{red}{6}8&textcolor{red}{F}0.tmp\System.dll

\end{Verbatim}
\newpage
\section{Hyperparameters}
\label{appendix:hyperparameters}

\subsection{Encoder-Decoder Hyperparameters}
\label{appendix:enc_dec_hyperparams}
We explore effects of latent representation size (Table \ref{tab:hp_representation_size}) and kernel width (Table \ref{tab:hp_kernel_width}) on the sample reconstruction quality. Using larger representation sizes increases reconstruction accuracy on the test set. Conversely, larger convolution kernels do not necessarily lead to better performance. The optimal choice of kernel width for our problem is 5.
\begin{table}[!h]
    \footnotesize
    \parbox{.45\linewidth}{
        \centering
        \begin{tabular}{cc}
        \toprule
        Representation Size & Reconstruction Accuracy \\
        \midrule
        32 & 91.06\% \\
        64 & 96.67\% \\
        128 & 98.26\% \\
        256 & 99.02\% \\
        512 & \textbf{99.28\%} \\
        \bottomrule
        \end{tabular}
        \caption{Effects of Representation Size}
        \label{tab:hp_representation_size}
    }
    \hfill
    \parbox{.45\linewidth}{
        \centering
        \begin{tabular}{cc}
        \toprule
        Kernel Width & Reconstruction Accuracy \\
        \midrule
        3 & 97.22\% \\
        5 & \textbf{98.50}\% \\
        7 & 98.48\% \\
        9 & 96.47\% \\
        \bottomrule
        \end{tabular}
        \caption{Effects of Kernel Width}
        \label{tab:hp_kernel_width}
    }
\end{table}

\subsection{Classifier Hyperparameters}
\label{appendix:clf_hyperparams}
We compare mean+max aggregation with learnable bag aggregator using different number of heads. Average classification accuracy and standard deviation are reported in Table \ref{tab:experiments:classifier_results}. Increasing the number of heads improves test classification accuracy, where $heads=64$ achieves better performance than the mean+max baseline.

\begin{table}[!ht]
\footnotesize
\centering
\begin{tabular}{lc}
\toprule
Model &           Accuracy \\
\midrule
Mean + Max    &  93.84\% ± 0.05\% \\
LBA(heads=2)  &  92.65\% ± 0.04\% \\
LBA(heads=4)  &  93.14\% ± 0.03\% \\
LBA(heads=8)  &  93.52\% ± 0.06\% \\
LBA(heads=16) &  93.85\% ± 0.07\% \\
LBA(heads=32) &  93.92\% ± 0.05\% \\
LBA(heads=64) &  \textbf{94.04\% ± 0.03\%} \\
\bottomrule
\end{tabular}
\caption{Classifier Accuracy}
\label{tab:experiments:classifier_results}
\end{table}

\end{document}